\documentclass[man]{apa6}
\usepackage{natbib}

\usepackage{graphicx, amssymb,subfigure,color,notes}
\usepackage[textwidth=2.5cm,textsize=tiny]{todonotes}
\newcommand{\hide}[1]{}

\title{Self-organizing maps and generalization: an algorithmic description of Numerosity and Variability effects}

\shorttitle{SOMs and Bayes}
\twoauthors{Valentina Gliozzi}{Kim Plunkett}
\twoaffiliations{Center for Logic, Language, and Cognition \&
Dipartimento di Informatica, 

\vspace{-0.15cm}
University of Torino 

\vspace{-0.15cm}
C.so Svizzera 185 - 10149 Torino, Italy

\vspace{-0.15cm}

 e-mail: valentina.gliozzi@unito.it}
{Department of Experimental Psychology, University of Oxford 

\vspace{-0.15cm}
 South Parks Road, Oxford OX1 3UD, UK
 
 \vspace{-0.15cm}
 e-mail: kim.plunkett@psy.ox.ac.uk}

\abstract{
Category, or property generalization is a central function in human cognition. It plays a crucial role in a variety of domains, such as learning, everyday reasoning,  specialized reasoning,  and decision making. Judging the content of a dish as edible, a hormone level as healthy, a building as belonging to the same architectural style as previously seen buildings, are examples of category generalization. In this paper, we propose self-organizing maps as candidates to explain the psychological mechanisms underlying category generalization. Self-organizing maps are psychologically and biologically plausible neural network models that learn after limited exposure to positive category examples,  without any need of contrastive information. Just like humans. They reproduce human behavior in category generalization, in particular for what concerns the well-known Numerosity and Variability effects, which are usually explained with Bayesian tools. Where category generalization is concerned, self-organizing maps are good candidates to bridge the gap between the computational level of analysis in Marr's hierarchy (where Bayesian models are situated) and the algorithmic level of analysis in which plausible mechanisms are described.
}

 
\begin{document}
\maketitle

\section{Introduction}
Category generalization is a central function in human cognition that plays a crucial role in  a variety of domains, such as learning, everyday reasoning, specialized reasoning, decision making. 
Judging the content of a dish as edible, a hormone level as healthy, a new building as belonging to the same architectural style than previously seen ones, are examples of category generalization. 

More formally, category generalization can be stated as follows. Starting from the observation that an object $x$ belongs to a category $C$, or that $x$ has the property $P$, how do we generalize $C$, or $P$, to objects other than $x$? 
As Shepard suggests: `Because any object or situation experienced by an individual is unlikely to recur in exactly the same form and context, psychology's first general law should be a law of generalization' \citep{shepard87}, p.1317.

Historically, probabilistic \citep{shepard87,anderson91,tengrif2001}, and similarity-based \citep{tversky77,medinschaffer1978,nosofsky86} accounts of generalization have faced each other: the former suggesting that our generalization judgements are based on probabilistic reasoning, the latter suggesting that indeed we base our generalization judgements on an evaluation of similarity between stimuli. 
So, for instance, according to the probabilistic account, the doctor's judgment on whether a new hormone level is healthy, similarly to a previously observed one, is based on a set of considerations involving the attribution of a probability to each possible range of values corresponding to healthy hormone levels, which in turn takes into account the likelyhood of encountering the observed examples of healthy hormone levels if {\em that} was indeed the correct range of possible hormone values. 
In contrast, according to the similarity-based approach, the doctor's judgement on whether a new hormone level is healthy is based on a completely different set of considerations based on the similarity of the new hormone level with respect to a previously built representation  (whether prototypical or exemplar-based or based on a pattern of neuron activations) of the category of healthy hormone levels. In the same family as similarity-based accounts can also be put the neural network accounts, in which generalization properties emerge from experience:  a correctly trained network will map a new stimulus to the category associated to the set of stimuli which are maximally similar to it \citep{gluckbower1988,kruschke1992,MareschalFrench2000,mcclellandrogers2003,guereckislove2004,alexnet}. 

In this paper we demonstrate that a specific kind of neural network, namely the self-organizing maps (SOMs for short) proposed by Kohonen (1982) can model human's generalization performance in a biologically and psychologically plausible manner. Unlike more traditional neural network models, SOMs reflect basic constraints of a plausible brain implementation. SOMs learn in an unsupervised fashion, i.e. by autonomously discovering relevant regularities in the training set, instead of requiring supervisory feedback needed in many other neural network models. They can learn from very few examples and training cycles, and without the need of negative examples or contrastive information, thus overcoming one of the main criticisms raised against neural network approaches \citep{GMHP2009}. Last, SOMs 
exhibit topological organization, in which similar stimuli are processed by close-by areas of the map, similarly to what happens in brain cortical areas \citep{kohonen2001,miikkulainen2005}.   
SOMs have proven to be successful models of category formation, capable of explaining experimental results, as well as of making novel and experimentally confirmed predictions \citep{schyns1991,miikkulainen2005,Li2007,GMHP2009,MayorPlunkett2010}.
In this paper we show that SOMs are adequate at reproducing two effects described by \cite{tengrif2001}, when they extend Shepard (1987)'s  universal law of generalization. Shepard describes an exponential decay in humans' inclination to consider a new stimulus as belonging to the same category as a previously considered one. The decay is proportional to the distance between the new stimulus and the examples already observed. 
Tenenbaum \& Griffiths extend Shepard's analysis to the case where the examples already observed are multiple. In this case, our inclination to generalize, and consider a new stimulus as belonging to the same category as the previously observed examples still decreases in an exponential manner when the distance between the new stimulus and the previously observed examples increases. This is affected by:
\hspace{5cm} 
\begin{itemize}
\item the {\em Numerosity Effect:} 
the more examples of a category observed within a given range, the lower
the generalization outside that range;

\item the {\em Variability Effect:}  the
higher the variability in the set of observed examples of a category, the
higher the generalization outside the examples' range.
\end{itemize}

We will show that SOMs exhibit both Numerosity and Variability Effects. Intuitively, the explanation of these effects lies in the fact that Numerosity augments the {\em accuracy} of category representation, with a consequent decrease in category generalization, whereas Variability decreases the accuracy of category representation, increasing the distance between the examples and the category representation, with a consequent increase in category generalization. 

To date, these effects are considered one of the main motivations supporting Bayesian analyses of category generalization, since Bayesian methods explain these phenomena in a very elegant way. It is also argued that alternative theories of categorization (ranging from exemplar and prototype theories to backpropagation-based neural networks) can not account for the same phenomena. Whence the conclusion that Bayesian models are best suited to explain category generalization \citep{tengrif2001}. 

However, Bayesian models of cognition are formulated at Marr's computational level of analysis, and therefore describe what an optimal solution to the problem of generalization would be. They do not describe a mechanism by which this solution could be found by a finite human, with limited capacities.

In this paper we show that, in contrast to the above claim, the same phenomena can be accounted for by SOMs, that embody notions such as representation and stimulus similarity that are well-established within the cognitive science literature, and that lie at the `algorithmic' level in Marr (1982)'s hierarchy. This explanation therefore  provides an account of the {\em psychologically plausible mechanisms} underlying categorization, and category generalization judgements, which is  lacking in Bayesian analyses of category generalization.

Establishing these mechanisms, and in general a bridge between Marr's different levels of analysis, is recognized as a key challenge by Bayesian cognitive scientists themselves, who have proposed some mechanisms that approximate Bayesian reasoning \citep{sanbornetal2010,shietal2010,sanbornchater2016}. The main problem of these proposed mechanisms is that they lack psychological plausibility, as we will discuss later. Given the promising results reported in this paper, we suggest that SOMs may be good candidates for describing psychologically plausible mechanisms that approximate Bayesian models of category generalization and of other related cognitive tasks. Even if understanding the precise relations between SOMs and Bayesian models needs further investigation, as it will become clear in the Discussion.

\section{Bayesian analyses of category generalization}

In contrast to more traditional models of cognition, which describe the psychological processes underlying cognitive abilities, Bayesian models of cognition are formulated at Marr (1982)'s level of `computational theory'. This means that in Bayesian models cognitive tasks are described as computational problems posed by the environment. Humans are assumed to find optimal solutions to these problems using inductive, probabilistic inference. By probabilistic inference, given some limited data, humans find the best possible hypothesis compatible with the evidence.

The problem of generalizing a category $C$ to a new object $y$ is formulated in this context as the problem of estimating the {\em probability} that $y$ belongs to $C$, after observing $X$ examples of $C$  \citep{shepard87,tengrif2001}\footnote{Shepard (1987) considers the case in which there is one single category example, whereas  Tenenbaum \& Griffiths (2001) extend the approach to multiple examples. In this paper we refer to Tenenbaum \& Griffiths (2001)'s theory.}. This goes through two steps.
 {\em First,} the {\em posterior probability} $p(h|X)$ is computed by the {\em Bayes Rule} for all possible extensions $h$ of $C$, where a possible extension is any consequential region, also called {\em hypothesis}.  The Bayes Rule uses both  priors and likelihoods. Each hypothesis $h$ in the space of hypotheses $H$ has a {\em prior} probability $p(h)$, independent from any observed example. Then there is the {\em likelihood} $p(X|h)$ of observing the examples $X$ if the true extension of the category was indeed $h$.  
The likelihood obeys the {\em size principle}: the smaller the size $|h|$ of the consequential region $h$ including all elements of $X$, the higher the probability of sampling all elements of $X$ as examples of $C$, and therefore the higher the likelihood. 

If  $X = \{x_i\}$ (with $1 \leq i \leq n$), then 

\begin{equation}
\label{likelyhood}
p(X|h) = \left\lbrace \begin{array}{l} \frac{1}{|h|^n} \quad{\rm if}\quad x_1 \dots x_n \in h;\\
 0 \quad{\rm otherwise.}\end{array}\right.   
\end{equation}

The priors and likelihood are combined together in order to determine the {\em posterior} probability of $h$ by the Bayes rule: 
\begin{equation}
\label{Bayes}
p(h|X)= \frac{p(X|h)p(h)}{\sum\limits_{h' in H}{p(X|h')p(h')}}
\end{equation}

{\em Second,} once the posterior probability $p(h|X)$ for all possible extensions $h$ has been computed by the Bayes Rule, the probability that $y$ belongs to $C$ is obtained by summing up the probability of all extensions containing $y$:

\begin{equation}
\label{probability that $y$ belongs to $C$}
p(y \in C | X)= \sum_{h:y\in h} p(h|X)
\end{equation}

In the hormone level example, if a doctor observes a {\em healthy} hormone level $x$, and she has to decide if another close-by hormone level $y$ is still healthy, she first has to infer a probability distribution over the set of possible extensions of the healthy hormone level. Only then she can estimate the probability that the new hormone level $y$ is still healthy.   

The computations just defined entail the {\em Numerosity} and the {\em Variability Effects} mentioned in the Introduction, that also hold in human category generalization \citep{tengrif2001}. 
By the {\em Numerosity Effect}, the more examples observed within a range, the lower the probability of generalization outside that range. This is due to the fact that by Equation \ref{likelyhood} the gap between the likelihood of smaller hypotheses and the likelihood of larger hypotheses increases exponentially with the number of examples observed, and by Equations \ref{Bayes} and \ref{probability that $y$ belongs to $C$} this gap is reproduced when generalizing outside the range of the examples. 
By the {\em Variability Effect},  the higher the variability in the set of observed examples, the higher the probability of generalization outside their range. 

The fact that Bayesian models of category generalization explain Numerosity and Variability effects in such an elegant way is one of the main motivations supporting Bayesian analyses of category generalization \citep{tengrif2001}. Indeed,  while the two effects also hold in human categorization, they are not easily captured by alternative theories of categorization (exemplar and prototype theories, backpropagation-based neural networks). In the next section we will argue that, on the contrary, a specific kind of psychologically plausible neural network, namely self-organizing maps, can account for these phenomena.  

For the moment, it is worth re-emphasising that Bayesian models of cognition are formulated at Marr's computational level of analysis, and do not describe a mechanism underlying these computations. Bayesian cognitive scientists recognize the urgency of establishing a bridge between Marr's different levels of analysis, and
acknowledge the identification of psychologically plausible processes underlying Bayesian inferences as a key challenge in their overall effort to reverse-engineer the human brain.
Some proposals have been made to provide a mechanistic account of the Bayesian models of generalization \citep{sanbornetal2010,shietal2010,sanbornchater2016}.
In particular, Sanborn et al. (2010) propose Monte Carlo methods as possible mechanisms underlying Bayesian models. Shi et al. (2010) propose mechanisms underlying Bayesian models that are based on exemplar models in which the stored exemplars correspond to hypotheses rather than stimuli. However, these models lack psychological realism: the tools they employ, both Monte Carlo methods and stored hypotheses, are far too complex to account as psychologically plausible mechanisms \citep{tengrif2001}. 

Relations between Bayesian approaches and neural networks have been studied in the past \citep{MacKay1995,Neal1996, McClelland1998}. However, the neural network models considered by these studies suffered from the criticism raised against neural networks as models on human category formation: in order to properly categorize and generalize these neural networks had to be trained with a lot of training examples (rather than few examples as humans), and with both positive and negative examples (whereas humans can learn from positive examples only).  

\section{Self-organizing map models of categorization}

In this section we show that a specific kind of neural network model, namely self-organizing maps, can provide a plausible description of the mechanisms underlying category generalization.  Self-organizing maps (SOMs for short) consist of a set of neurons, or units, spatially
organized in a  grid \citep{kohonen2001}.  

Each map unit $u$ is associated with a
weight vector $w_u$ of the same dimensionality as the input vectors.

At the beginning, all weight vectors are initialized to random values, outside the range of values of the input stimuli. 
 
During training, the input elements  are sequentially presented to all neurons of the map. After each presentation of an input $x$, the {\em best-matching unit} (BMU$_x$) is selected: this is the unit $i$ whose weight vector $w_i$ is closest to the stimulus $x$ (i.e. $i = \arg\min_j\|x - w_j\|$).

The weights of the best matching unit and of its surrounding units are updated in order to maximize the chances that in the future the same unit (or the surrounding units) will be selected as the best matching unit for the same stimulus or for similar stimuli. At iteration $n+1$, the weights for neuron $j$ are updated as follows:

\begin{equation}
\label{eq:som_update_rule}
w_j(n+1) = w_j(n) + \eta(n) h_{BMU_x,j}(n) (x - w_j(n))
\end{equation}

\noindent where $\eta$ is the \emph{learning rate}, and $h_{BMU_x,j}$ is the neighborhood function between the best-matching unit $BMU_x$ and $j$.  $h_{BMU_x,j}(n)$ is defined as $h_{BMU_x,j}(n) = exp^{\frac{-d_{BMU_x,j}^2}{2\sigma(n)^2}}$, where $d_{BMU_x,j}$ is the distance between $BMU_x$ and $j$ on the map's grid, and $\sigma(n)$ is the width of the gaussian.

This weight change has a twofold effect:
\begin{itemize}
\item it reduces the distance between the Best-matching unit (and its surrounding neurons) and the incoming input, so that in the future that same unit (and the surrounding ones) will be most likely to be the best matching unit for the same or similar inputs;
\item  it organizes the map topologically so that the weights of close-by neurons are updated in a similar direction, and come to react to similar inputs. 
\end{itemize} 

In our simulations we have used a 3*3 hexagonal SOM trained with learning rate $0.5$, with neighborhood a gaussian with width $0.5$. Each input stimulus was presented to the SOM {\em once} during learning\footnote{The parameters chosen allow to  present to the map the training set few times, in this case only once. Similar results hold for larger SOMs, with standard parameters settings, i.e. lower learning rate and hundreds of training epochs.}. With this training schedule SOMs learn without extensive training, and from positive examples only.

The learning process is incremental: after the presentation of each input, the map's representation of the input (and in particular the representation of its best-matching unit) is updated in order to take into account the new incoming stimulus.  This can be seen as a plausible mechanism by which humans form categories: starting from a first stimulus, that gives rise to a first starting representation, the representation is updated each time a new stimulus is considered, in order to accommodate it. The final representation of the stimuli is the result of this iterative process. 
At the end of the whole process, the SOM has learned to organize the stimuli in a topologically significant way: similar inputs (with respect to Euclidean distance) are mapped to close areas in the map, whereas inputs which are far apart from each other are mapped to distant areas of the map.

\subsection{Numerosity and variability effects within self-organizing maps}

Do numerosity and variability have an effect on category formation and generalization within SOMs?

In order to answer the question, we have run three simulations to investigate how the SOMs' organization of the input stimuli was affected by numerosity and variability. In each simulation we have trained the network with a different set of stimuli. Stimuli are points in a continuous metric psychological space (e.g., hormone levels) that  vary along one dimension (as for Tenenbaum \& Griffiths, 2001). In the first simulation, the Base Condition, we trained the map with two stimuli: the points 
$[50, 0]$ and $[60, 0]$. In the second simulation we have augmented the number of stimuli used for training, by keeping the range of values the same as in our Base Condition. The training stimuli in this simulation, called {\em Numerosity Condition} are the points: $[50,0], [53,0], [55,0], [57,0], [59,0], [60,0]$.
In the third simulation we have kept the number of stimuli presented to the SOM for training constant with respect to the Base Condition, while varying the range of values. In this simulation, our {\em Variability Condition}, the stimuli considered are $[30,0]$ and $[60,0]$\footnote{The stimuli vary in one direction to use the same stimuli as Tenenbaum \& Griffiths, 2001. The results would hold also when considering stimuli varying in both directions.}.

After learning  is complete, we focus  on the category representation formed by the map out of the examples.
We take a minimalist notion of what is the {\em map's category representation}: this is the ensemble of best-matching units corresponding to the examples of the category \footnote{We could add to the map's {\em category representation} the neurons that are very close to the best-matching unit, both on the map's grid and on the input space. The results still hold.}.
We will use $BMU_{C}$ to refer to the map's representation of category $C$.

To start with, note that both the Variability and the Numerosity Condition affect the {\em accuracy} by which the category examples are represented by the SOM. In the Numerosity Condition the examples are represented more accurately than in the Base Condition: the distance between the map's category representation and the examples is lower in the Numerosity Condition than in the Base Condition (Fig. $1$). 
In contrast, variability leads to a less accurate representation of the examples: in the Variability Condition the distance between the map's category representation and the category examples is higher than in the Base Condition (Fig. $1$). 

\begin{figure}[htbp]
\vspace{-4cm}
\includegraphics[width=\textwidth]{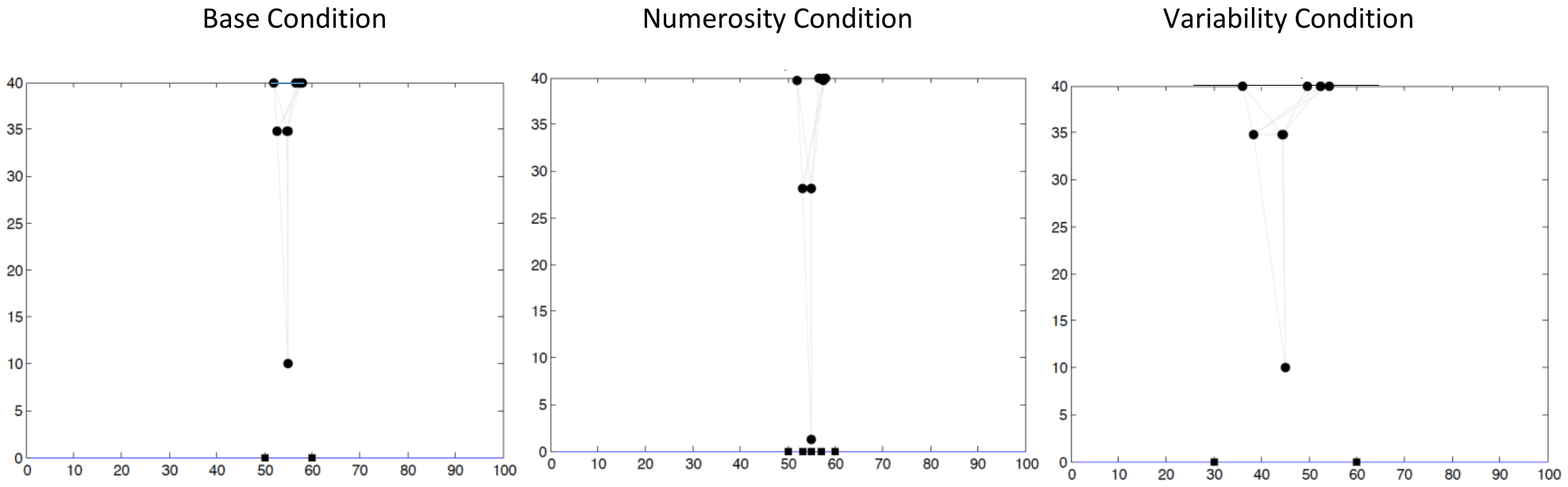}\vspace{-12cm}\caption{SOM organization in the Base Condition, in the Numerosity Condition, and in the Variability Condition. Each plot has nine black dots, one for each neuron of the SOM (plotted with respect to its weights' values, on the $x-$axis the first value, on the
$y-$axis the second value). The grey lines connect adjacent neurons in the map's grid. Each plot represents the neurons' organization after training with the stimuli in the Base, Numerosity, and Variability Condition, respectively. Black boxes represent the stimuli (plotted with the first value on the $x-$axis, the second value on the $y-$axis) in the three conditions. Consider the stimuli best-matching unit. In all three conditions this is the map's bottom unit. The maximal Euclidean distance between the best-matching unit and the stimuli is higher in the Base Condition than in the Numerosity Condition, whereas it is lower in the Base Condition than in the Variability Condition.
}\label{figtest}
\vspace{1cm}
\end{figure}

{\em Does this difference in the accuracy of representation of the examples affect the pattern of generalization of category membership to new stimuli?}

The answer is positive.
The accuracy of representation of the category examples sets up a {\em tolerance level}, that is then used when making a generalization judgment.   This tolerance level is determined by the discrepancy between the model's internal category representation and the new stimulus.so-called {\em quantization error} which is a common and useful measure of the quality of a map's representation: for each example $x$ this is the Euclidean distance of $x$ from its closest neuron( $\|x - w_{BMU_{x}}\|$).
This tolerance level, in turn, has a role when judging the {\em significance} of the distance of a new stimulus with respect to the category representation. A given distance of a new stimulus $y$ from a category representation will seem more significant if our tolerance level is low: in this case any deviation from the category representation will be considered significant. In case our tolerance level is high we will be more tolerant to the same deviation from category representation, and judge the same distance less significant. 
This is formally captured by the notion of {\em relative distance} of a new stimulus $y$ with respect to the representation of a category $C$ ($BMU_{C}$): this is the distance of $y$ with respect to the category representation \footnote{We take the distance from the {\em closest} neuron within the map's category representation}, compared with the distance from the same representation of the category examples\footnote{We take the {\em maximal} such error.}. 

\begin{equation}
\label{relative-distance}
\frac{\min \|y - BMU_{C}\| }{\max_{x \in C} \| x - BMU_{x}\| }
\end{equation}

Using Equation \ref{relative-distance} we can define the  {\em map's generalization degree} of category $C$ membership to a new stimulus $y$ as the inverse of the distance of  $y$ from the category representation.

This  measure of the map's disposition to generalize leads to both Numerosity and Variability effects. When the number of category examples within a given range increases, the generalization curve outside that range shrinks, and the SOM is less likely to attribute to the same category a new stimulus $y$ outside that range (Fig. $2$). 

The opposite effect holds when the variability of the category examples increases: in this case the generalization curve widens, and the SOM's disposition to attribute a new stimulus $y$ to the same category increases (Fig. $2$). 
\begin{figure}[htbp]
\vspace{-3cm}
\includegraphics[width=\textwidth]{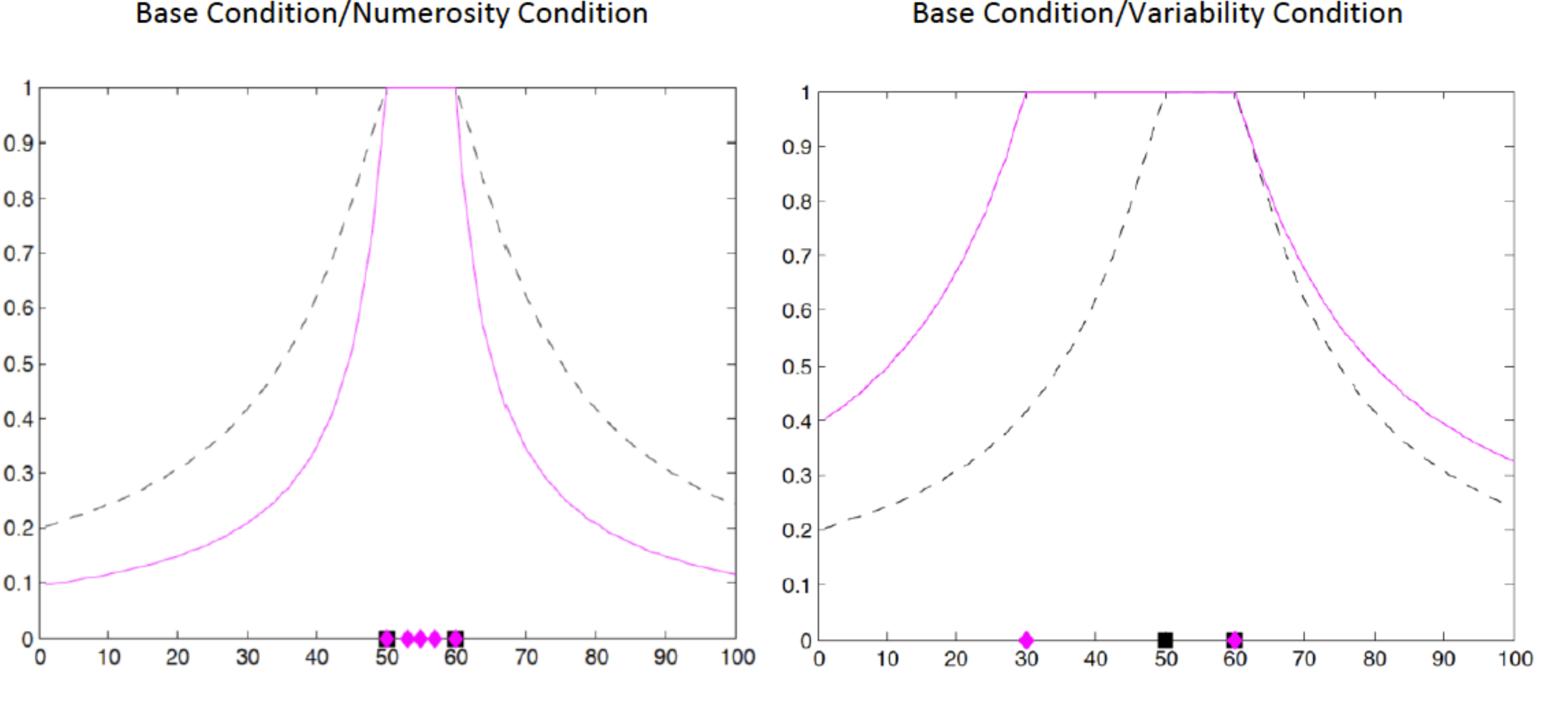}\caption{Map's  generalization. On the $x-$axis the stimuli varying along that dimension. On the $y-$axis the corresponding {\em relative-distance} (Equation \ref{relative-distance}) from the map's category representation. Left plot: black dashed curve for the {\em relative-distance} in the Base Condition (where black squares are the examples for this condition), magenta continuous line for the {\em relative-distance} in the Numerosity Condition (where magenta diamonds are the examples for this condition). The  curve in the Numerosity Condition has lower values than the Base Condition curve, indicating lower generalization levels. Right plot: Black dashed curve for the {\em relative-distance} in the Base Condition, magenta continuous line for the {\em relative-distance} in the Numerosity Condition.  The  curve in the Variability Condition has higher values than the Base Condition curve, indicating higher generalization levels.}

\end{figure}

\section{Discussion}

The results of the previous section show that SOMs can provide a mechanistic account of the Numerosity Effect as well as of the Variability Effect observed in human categorization. In contrast to the explanation advocated on a Bayesian account, the SOM's explanation of the two effects relies only on a notion of distance of the new stimulus from the category representation. As far as we are aware, this is the first quantified argument demonstrating that the two effects, which characterize human category generalization, can be explained within the similarity-based paradigm. Furthermore, SOMs exhibit the two effects when exposed only to few positive category examples (as in humans), without the need of extensive or contrastive learning.  SOMs results easily extend to stimuli that vary along more than one dimension, as long as the notion of Euclidean Distance between stimuli can be clearly defined.

These effects cannot be explained within 
traditional theories of categorization based on similarity, such as the prototype theory \citep{PosnerKeele1968} or the exemplar theory \citep{medinschaffer1978,nosofsky86}. Indeed, for none of these theories does numerosity play a role: the prototype remains the same independently from the number of instances considered, and the distance of a new stimulus from the set of exemplars does not change when the number  of exemplars is augmented.  In both theories variability leads to a shift of the generalization curve only in the direction of variability, but there is no general widening of the generalization curve.  

Numerosity and Variability effects cannot be explained within neural networks based on backpropagation either, since these networks need contrastive information in order to achieve a reasonable categorization of the inputs, whereas in the examples considered here there are only few positive instances of the category.

Within the exemplar models or combinations of exemplar and neural network models or related models \citep{kruschke1992,sustain}, these effects can be tailored by deliberately letting the attentional weights or the receptive fields change as a consequence of numerosity or variability of the examples.  But this is an external, ad hoc explanation. On the contrary, within SOMs the two effects emerge from the SOM learning algorithm, with no external, ad hoc ingredient. 

{\em Could it be that SOMs are an implementation at the algorithmic level of the Bayesian analyses of category generalization?} 

It is very difficult to answer this question analytically. This paper shows that indeed SOMs capture {\em some} aspects of Bayesian analyses. Being psychologically and biologically plausible, this makes SOMs good candidates to bridge the computational, Bayesian level of analysis and the algorithmic level of analysis in the study of category generalization (and purportedly for other cognitive tasks). However, understanding the exact extension of this correspondence requires future research.

 \begin{figure}
 \vspace{-2cm}
\includegraphics[width=\textwidth]{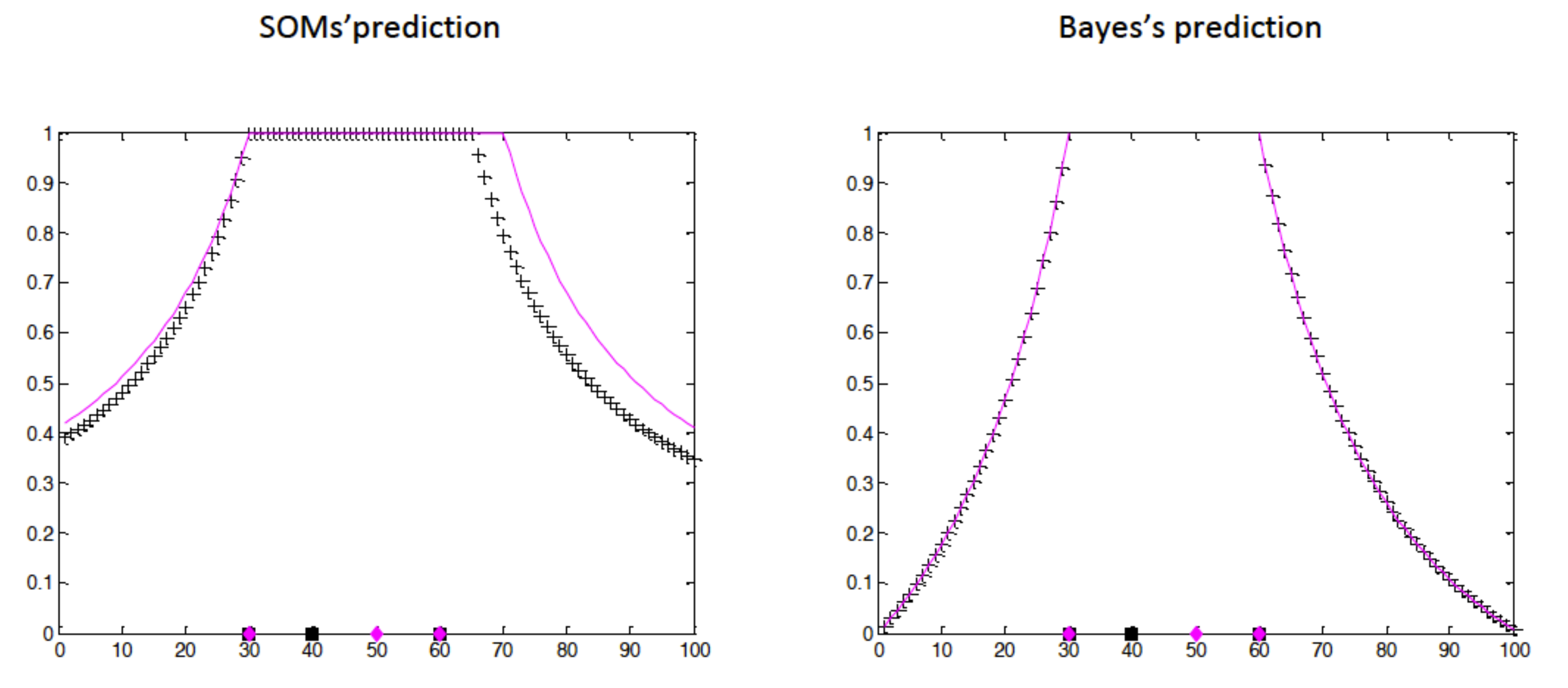}\caption{On the $x-$axes the stimuli varying along that dimension. Black boxes are the examples of Set $1$, magenta diamonds are the examples of Set $2$. Left plot: on the $y-$axes the relative-distance (Equation \ref{relative-distance}) of the stimuli from the map's representation of Set $1$ (black) and from the map's representation of Set $2$ (magenta), respectively. Right plot: Probability (Equation \ref{probability that $y$ belongs to $C$}) that the corresponding stimuli belong to the same category than the observed examples of Set $1$ (black)  and Set $2$ (magenta), respectively.}
\end{figure}

For the time being we can say that there are small differences between SOMs and Bayesian analyses predictions. For instance, SOMs are sensitive to the specific position of repeated category examples within a given range, whereas Bayesian analyses are not (at least when the size principle is used as a likelihood estimation, as in Tenenbaum and Griffiths, 2001). Take  the two following sets of category examples, whose values vary in the same range but in which the exact values of the instances change: {\bf Set 1} $=\{[30,0],[40,0],[60,0]\}$;  {\bf Set 2}$=\{ [30,0],[50,0],[60,0]\}$.

Figure $3$ shows that for the Bayesian analysis of categorization in the two conditions the generalization curve will be the same \citep{tengrif2001}. On the contrary, for SOMs there will be a difference in the generalization curve in the two conditions. This is the consequence of the fact that SOMs form a representation of the examples, whose position is shifted depending on the exact values of the category examples. Instead, in Bayesian models there is no representation being formed, therefore no shift, and no consequent effect of the exact values of category examples.
 
\section{Conclusions}
In this paper we have shown that a biologically and psychologically plausible neural network architecture can provide a mechanistic account of Numerosity and Variability effects usually explained, at the computational levels, with Bayesian tools. SOMs can do so when exposed to limited category examples without any need of contrastive information, thus contradicting the main criticism against models of category generalization alternative to the Bayesian ones. We leave for future research the investigation of the extension of the correspondence between SOMs and Bayesian models, and whether SOMs can be seen as describing the mechanisms underlying Bayesian analyses in general.


\bibliographystyle{apa}
\bibliography{CognitiveScienceJournal2017}

\end{document}